\documentclass[10pt,twocolumn,letterpaper]{article}

\usepackage{cvpr}              

\usepackage[dvipsnames]{xcolor}
\usepackage{adjustbox}
\usepackage[algo2e,ruled,linesnumbered,vlined]{algorithm2e}
\usepackage{microtype}
\usepackage{multirow}

\definecolor{cvprblue}{rgb}{0.21,0.49,0.74}
\usepackage[pagebackref,breaklinks,colorlinks,citecolor=cvprblue]{hyperref}

\def\paperID{68_13} 
\def\confName{CVPR}
\def\confYear{2024}

\title{Dataset condensation with latent quantile matching}

\author{Wei Wei$^{1}$, Tom De Schepper$^{1}$, Kevin Mets$^{2}$\\
University of Antwerp - imec, IDLab, \\$^{1}$Department of Computer Science, $^{2}$Faculty of Applied Engineering,\\ Sint-Pietersvliet 7, 2000 Antwerp, Belgium\\
{\tt\small \{wei.wei, tom.deschepper, kevin.mets\}@uantwerpen.be}
}

\begin{document}
\maketitle
\begin{abstract}
Dataset condensation (DC) methods aim to learn a smaller, synthesized dataset with informative data records to accelerate the training of machine learning models. Current distribution matching (DM) based DC methods learn a synthesized dataset by matching the mean of the latent embeddings between the synthetic and the real dataset. However, two distributions with the same mean can still be vastly different. In this work, we demonstrate the shortcomings of using Maximum Mean Discrepancy to match latent distributions, i.e., the weak matching power and lack of outlier regularization. To alleviate these shortcomings, we propose our new method: Latent Quantile Matching (LQM), which matches the quantiles of the latent embeddings to minimize the goodness of fit test statistic between two distributions. Empirical experiments on both image and graph-structured datasets show that LQM matches or outperforms previous state of the art in distribution matching based DC. Moreover, we show that LQM improves the performance in continual graph learning (CGL) setting, where memory efficiency and privacy can be important. Our work sheds light on the application of DM based DC for CGL.
\end{abstract}    
\begin{figure*}[h]
    \subfloat[Synthetic latent feature generated using MMD as distance function. The mean of the distribution matches. However, the synthetic distribution contains an outlier and there are no values in between 0.5 and 1.5. MMD fails to approximate the distribution.]{\makebox[0.48\textwidth][c]{\includegraphics[width=0.33\textwidth]{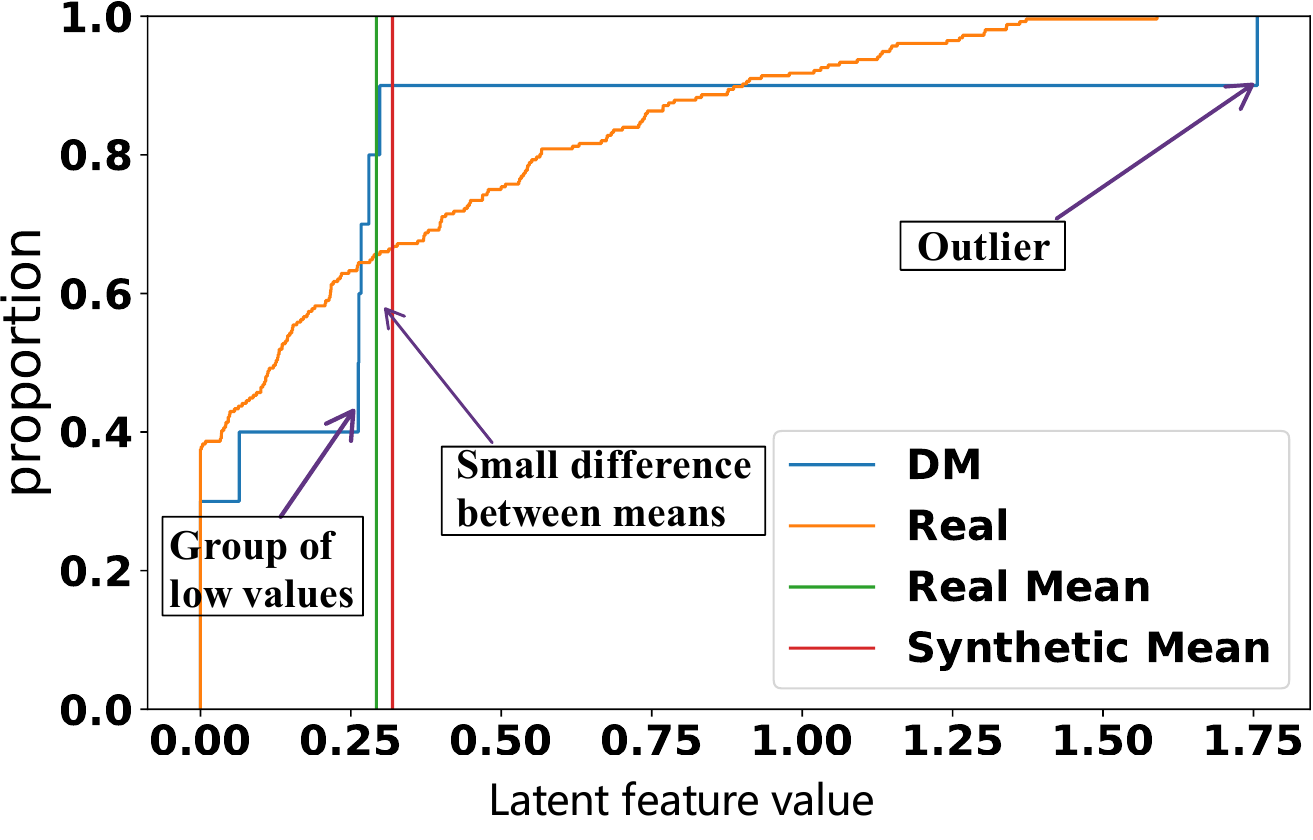}}\label{fig:kde_MMD}} 
\hfill
    \subfloat[Synthetic latent feature generated using LQM as distance function. LQM minimizes the distance between synthetic data records and the optimal k-point approximations of the real distribution. LQM approximates the distribution better.]{\makebox[0.48\textwidth][c]{\includegraphics[width=0.32\textwidth]{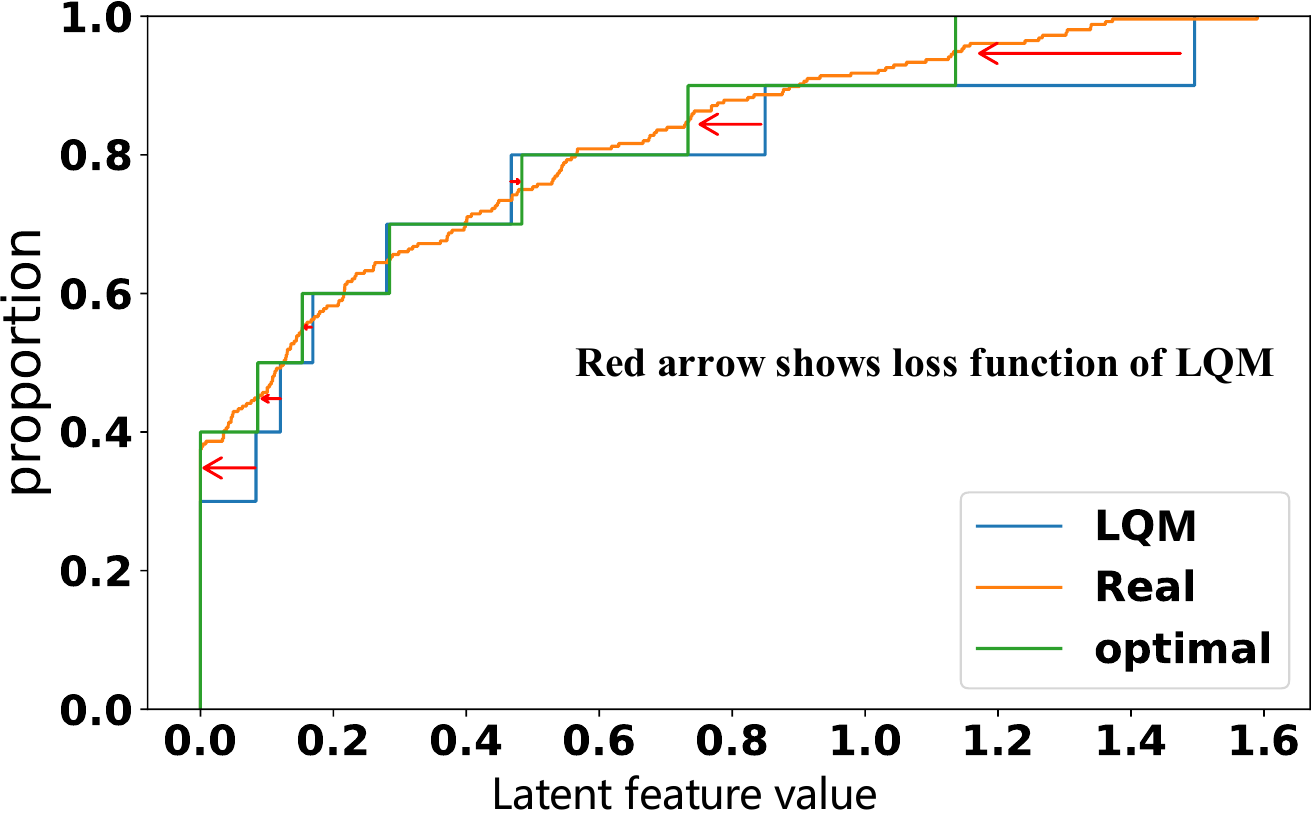}}\label{fig:kde_LQM}} 
\centering
\captionsetup{justification=centering,margin=0cm}
\caption{The empirical cumulative distribution function (ECDF) of a latent feature of class 0 in CIFAR-10 after 1900 epochs of training.}
\label{fig:teaser}
\end{figure*}

\section{\uppercase{Introduction}}
\label{sec:introduction}
As the world becomes more connected, the available data increases exponentially. In order to learn knowledge that is generalizable to all the available data, large and complex deep models are proposed. The immense computational cost of training these models hinders their development, which is undesirable. Next, real-world datasets may contain sensitive data. These data can't be made public due to the privacy concerns. This hinders the research transparency and reproducibility. Dataset condensation (DC) can solve these issues by reducing the size of the training set. Instead of a large training set with potential redundant data records, DC methods aim to generate a small, synthetic dataset that is highly informative, which is also shown to be more privacy-preserving \cite{dong2022privacy}. The deep learning models trained on the synthetic dataset should have similar evaluation performance compared to the models trained on the original, large training set. DC is applied to many problems, such as continual learning (CL) \cite{liu2023catcgl} \cite{yang2023efficient}, federated learning \cite{goetz2020federated} and neural architecture search \cite{such2020generative}. DC improves the state-of-the-art (SOTA) performance of CL with graph-structured data significantly \cite{liu2023catcgl}. However, the studies on applying DC to continual graph learning (CGL) problems are still scarce \cite{liu2023catcgl}. Our work focuses on improving DC and its application on CGL, we hope that our work sheds light on this underexplored field. We use the terms real dataset and original dataset interchangeably to refer to the dataset to condense.

Based on \citet{sachdeva2023data}, the SOTA dataset condensation methods are categorized into four categories: 1) Meta-Model Matching, 2) Gradient Matching, 3) Trajectory Matching, 4) Distribution Matching. The former three categories involve expensive bi-level optimization to achieve the goal of DC. I.e., they train a model for each optimization step of the synthetic dataset. Our work focuses on the last category, the distribution matching (DM) based DC methods. These methods do not have the costly bi-level optimization. Instead of comparing the distance between the learned parameters, or the gradients of the trained models, DM based DC methods initialize different models for each optimization steps, extract latent embeddings of both real and the synthetic dataset, and matches their distribution.

SOTA DM based DC methods use the Maximum Mean Discrepancy (MMD) to measure the distance between synthetic and real latent embeddings \citep{zhao2023datasetdm,zhao2023improvedidm,wang2022cafe}. We note that comparing the mean of the latent embeddings is equivalent to comparing the sum of the mean of each individual latent feature distribution in the latent embedding. In the remainder of this work, we focus on the comparison between individual latent feature distributions. MMD only matches the mean of these distributions (i.e., first-order moment). As it does not provide any guarantee about other aspects of the matched distributions, they can still differ in statistics such as the variance, skewness, and kurtosis. This issue was identified by \citet{zhao2023improvedidm}. However, the author still used MMD as the distance function despite this shortcoming. As an empirical evidence, an example where the MMD fails to match the latent distribution is shown in \cref{fig:kde_MMD}. In the synthetic dataset generated by MMD, there is one sample with very high value, while other samples have relatively low values. This matches the mean correctly, but it does not reflect the real distribution. The synthetic feature with the highest value is also larger than the maximum value in the real dataset, this may harm the learning process of the model.

Based on this observation, we believe that replacing MMD with a more suitable metric that reflects the distance between the distributions better, can improve the performance of DM based DC methods. We are inspired by goodness of fit tests in the field of statistics, which quantifies a distance between two empirical cumulative distribution functions (ECDF). Specifically, we propose a novel DM based DC method: Latent Quantile Matching (LQM) based on the two-sample Cramér-von Mises (CvM) test \cite{anderson1962distributioncramervonmises}. It estimates the squared difference between the two ECDFs. It is empirically shown to be more powerful \cite{stephens1974edf} than the widely used Kolmogorov-Smirnov test \cite{massey1951kolmogorov}, and there is a simple way to minimize the CvM test statistics for any number of data records $k$ based on the optimal $k$-point discrete approximation studied in \citet{kennan2006note}.

The objective of LQM is to minimize the difference between synthetic data records and optimal quantiles $k$ in the latent space, which corresponds to the optimal $k$-point discrete approximation of the original latent distribution. This objective is similar to the Quantile Regression (QR) problem in the field of statistics \cite{koenker2001quantile}. However, the objective of QR is to express the quantiles of the distribution of the response variables as functions of the observable input variable. Our approach does not explicitly learn a model which predicts the values at a certain quantile. Instead, LQM tries to learn a small synthetic dataset that matches the different quantiles of the real dataset in the different latent spaces extracted by the feature extractors. Our approach tries to minimize the difference between synthetic latent features and the quantiles of the real latent features. In \cref{fig:kde_LQM}, we show that the objective function is defined by the horizontal difference between the blue ECDF and the optimal green ECDF. As a result, the synthetic dataset produced by LQM discourages any outliers and approximates the real distribution better. Our contributions are as follows.

\begin{itemize}
\item We demonstrate the shortcomings of Maximum Mean Discrepancy as a metric to measure the distance between two empirical distributions. We reveal that two distribution with the same mean may not be similar.

\item We propose a novel distribution matching based dataset condensation method: Latent Quantile Matching. We learn a small set of synthetic samples which match the different quantiles of the original dataset in the latent space. The quantiles are chosen based on the optimal k-point discrete approximation of the original latent distribution \cite{kennan2006note}.

\item We extensively evaluate LQM on image and graph datasets. 
Compared to previous studies, our experiments show the efficacy of our method with different data structure. Next, the experimental results show that the model trained on dataset learned by LQM outperform the models trained on dataset learned by MMD on most datasets. I.e., the performance gap between the model trained on the smaller synthetic dataset and the full, original dataset is reduced.

\item We evaluate LQM in continual graph learning (CGL) setting with CaT-CGL \cite{liu2023catcgl}. We show that LQM outperforms SOTA CGL method. The performance improvement is larger when the synthetic dataset is small.
\end{itemize}

\section{\uppercase{Related works}}\label{section:relatedworks}
\textbf{Dataset Condensation.} Dataset condensation (DC) aims to condense large training datasets into small synthetic datasets, while preserving the same evaluation performance when the model is trained only on the synthetic datasets. \citet{wang2018datasetdistillation} proposed a way to maximize the performance of model trained using the synthetic dataset by using meta-learning. Subsequent studies introduced different techniques such as soft-label \cite{bohdal2020flexiblelabel}, gradient-matching \cite{zhao2020datasetgradientmatching}, data-augmentation \cite{zhao2021datasetaugmentation}, trajectory matching \cite{cazenavette2022datasettrajectory} and data-parameterization \cite{kim2022datasetparameterization}. Despite their good performance, most of the methods rely on bi-level optimization involving second-order derivatives, which is computationally expensive.

On the contrary, Distribution Matching (DM) \cite{zhao2023datasetdm} proposes to condense the large training dataset into a synthetic dataset by matching the latent distribution between the two datasets using different, randomly initialized feature extractors. The distance between the latent distributions is measured by the Maximum Mean Discrepancy (MMD). DM does not require bi-level optimizations and is less computationally expensive compared to other approaches. Subsequent studies outperform DM by introducing techniques such as partition and expansion augmentation \cite{zhao2023improvedidm}, trained feature extractors \cite{zhao2023improvedidm} and attention matching \cite{sajedi2023datadamattention}. Despite the improvement in performance, all the above-mentioned methods use MMD to measure the distance between the two distributions. This makes them vulnerable to the shortcoming shown in \cref{fig:kde_MMD}. In this work, we replace MMD with a distance function inspired by the goodness of fit tests in the field of statistics. Our distance function only reaches zero if the synthetic latent distribution is close to the specified quantiles of the original latent distribution for all classes. 

Recently, \citet{liu2023datasetwasserstein} also studied the shortcomings of MMD, and replaces it with more powerful metrics. Our work is concurrent with theirs. Their work is based on optimal transport theory and uses Wasserstein Barycenter \cite{agueh2011barycenters} as the matching metric. The computation of Wasserstein Barycenter is computationally expensive \cite{borgwardt2021computational}, and they used an approximation algorithm proposed by \citet{cuturi2014fast} to reduce it. In contrary, our work is not computationally expensive as it is based on statistical goodness of fit tests and uses the mean squared errors to the target quantiles as the metric. The computation of quantiles of an empirical distribution is not computationally expensive. By sorting the empirically drawn samples, we can retrieve all the quantiles of an ECDF. As an example, the time complexity of quicksort is $O(nlog(n))$ \cite{hoare1962quicksort}.

\textbf{Goodness of Fit Tests.} In statistics, a goodness of fit test measures how well a set of observed values fits a given statistical model. A subset of the tests quantifies the distance between two ECDFs. The Kolmogorov–Smirnov (KS) test \cite{massey1951kolmogorov} measures the maximum difference between the two ECDFs. The Cramér–von Mises (CvM) test \cite{anderson1962distributioncramervonmises} estimates the squared differences between the two ECDFs. The Anderson-Darling (AD) test \cite{scholz1987ksampleandersondarling} extends the CvM test by assigning more weights to the tails of the distributions. Previous studies show that the CvM and AD test are more powerful than the KS test \cite{razali2011powercomparison}. Compared to MMD, these goodness of fit tests can match distributions beyond the first-moment. Our work uses optimal quantiles that minimize the CvM test statistic (CvM stat), proved by \citet{kennan2006note} and reported in \citet{barbiero2023discrete}, as the regression target for the synthetic dataset. This ensures the similarity of the synthetic and the real latent distributions.

\textbf{Continual (Graph) Learning.} Continual Learning (CL) is a research field that benefits from DC. CL aims to build deep models that can acquire knowledge across different tasks without retraining from scratch. Amongst different CL methods, replay based methods obtain exceptional performances \cite{prabhu2020gdumb} \cite{zhang2022cglb} \cite{bagus2021investigation} \cite{wei2024benchmarking}. They store a fraction of the data from the past tasks as memory buffer and replay them during the training of the new tasks. However, they require additional storage space and may raise privacy concerns. The advancements in dataset condensation show a promising direction to solve these problems. By condensing the dataset, the storage overhead can be reduced. Next, the privacy issue is alleviated, as the synthetic dataset is an obscured version of the real dataset, which is shown to be safe against membership-inference privacy attacks \cite{dong2022privacy}. To the best of our knowledge, CaT-CGL \cite{liu2023catcgl} is the only work that studies DC in Continual Graph Learning (CGL). We incorporate our method into CaT-CGL and outperforms their method.

\section{\uppercase{Preliminaries}}\label{section:preliminaries}
\textbf{Dataset Condensation.} Given a large training set $\mathcal{T} = {(x_1,y_1),...,(x_{|\mathcal{T}|},y_{|\mathcal{T}|})}$ containing $|\mathcal{T}|$ data records and their labels, DC aims to synthesize a smaller set $\mathcal{S} = {(s_1,y_1^s),...,(s_{|\mathcal{T}|},y_{|\mathcal{T}|}^s)}$, $|\mathcal{S}|\ll|\mathcal{T}|$, such that the deep models trained on $\mathcal{S}$ maintain similar evaluation performance as models trained on $\mathcal{T}$. Let $x$ be a sample from the real data distribution $P_D$ with label $y$, $\phi_{\theta^\mathcal{T}}$ and $\phi_{\theta^\mathcal{S}}$ be two variants of the same model $\phi$ with parameters $\theta$ trained on dataset $\mathcal{T}$ and $\mathcal{S}$, respectively, and $\mathcal{L}$ denotes the loss function (e.g., cross-entropy loss). The objective of DC is to find the ideal dataset $\mathcal{S^*}$ defined by the following equation.
\begin{equation}\label{eqn:DC}
\resizebox{0.91\columnwidth}{!}{
$\mathcal{S^*} = \underset{S}{argmin}\,\mathbb{E}_{x\sim P_{D}}||\mathcal{L}(\phi_{\theta^\mathcal{T}}(x),y)-\mathcal{L}(\phi_{\theta^\mathcal{S}}(x),y)||$
}
\end{equation}
\textbf{Distribution Matching.} Distribution Matching (DM) \cite{zhao2023datasetdm} proposed to tackle the DC problem by matching the distributions of the latent embeddings $\phi_{\theta}(x_i)$ and $\phi_{\theta}(s_j)$ for $\mathcal{T}$ and $\mathcal{S}$, respectively. The objective of distribution matching based DC methods is to find the ideal dataset $\mathcal{S}^*$ such that:
\begin{equation}\label{eqn:DM}
\mathcal{S^*} = \underset{S}{argmin}\,\mathcal{D}(\phi_{\theta}(\mathcal{T}),\phi_{\theta}(\mathcal{S}))
\end{equation}
Where $\mathcal{D}$ denotes a distance function which takes as input the latent embeddings extracted by randomly initialized networks with parameter $\theta\sim P_{\theta_{0}}$. Compared to the general objective of DC, described in \cref{eqn:DC}, DM's objective function does not require the model to be trained. Instead, DM attempts to learn the synthetic dataset with the same distribution as the real dataset with randomly initialized networks. This surrogate objective makes the computational cost of DM based DC methods lower than the approaches that require bi-level optimization. When the distance function $\mathcal{D}$ is MMD, the objective is defined as:
\begin{equation}
\mathcal{S^*} = \underset{S}{arg\,min}\,\mathbb{E}_{\theta\sim P_{\theta_{0}}}||\frac{1}{|\mathcal{T}|}\underset{i=1}{\sum^{|\mathcal{T}|}}\phi_{\theta}(x_i) - \underset{j=1}{\sum^{|\mathcal{S}|}}\phi_{\theta}(s_j)||
\end{equation}
\textbf{Continual (Graph) Learning.} The two most used settings in CL are Task-incremental (Task-IL) and Class-incremental (Class-IL). We focus on the more difficult Class-IL setting \cite{zhou2023deep}. We assume a sequence of $B$ tasks: $\{D^1,D^2,D^3,...D^B\}$. $D^b = \{(x^b_i,y^b_i)\}_{i=1}^{n_b}$ denotes the $b$-th incremental training task with $n_b$ data instances. Each task $D$ contains data instances from one or more classes $c \in C$. We denote function $Cls$ as a function that maps the task to the classes of the data instances it contains. The classes in each task $D$ do not overlap with any other tasks defined in the sequence. Thus:
\begin{equation}
Cls(D^a) \cap Cls(D^b) = \emptyset, \forall a,b \in \{1,...,B\}, a\neq b    
\end{equation}
During the $b$-th incremental training process, the model only has access to the data instances from the current task $D^b$. Rehearsal based methods can still access their memory buffer. After each incremental training process, the model is evaluated over all seen classes $C_{seen}^b = Cls(D^1) \cup ... \cup Cls(D^b)$. Class-IL aims to find the model $f(x):X \mapsto C$ that minimizes the loss over all the tasks in the sequence.


\section{\uppercase{Methodology}}\label{section:methodology}
The base of our method lies on the two shortcomings that we observed with dataset condensation methods that use Maximum Mean Discrepancy (MMD) to measure the distance between the synthetic and the real latent embeddings. In this section, we explain the two shortcomings in detail, show that they can harm the distribution matching objective, and propose our new method: Latent Quantile Matching that alleviates the shortcomings.

\subsection{Shortcomings of MMD}\label{sec:shortcomings}
\textbf{Inadequate distribution matching power. } As the name Maximum Mean Discrepancy suggests, MMD only matches the mean of the empirical distributions (i.e., first-order moment). A previous study \cite{gretton2012kernel} has shown that, in the universal reproducing kernel Hilbert space (RKHS), asymptotically, MMD is 0 if and only if $P_X = P_Y$. Characterizing the function space explored by neural networks is a difficult and ongoing field of study. Recent works show that the function space of fully connected layers can be defined as trainable ladders of Hilbert spaces \cite{chen2023multirkhs}, or as reproducing kernel Banach space \cite{bartolucci2023understandingrkbs}. However, there are no studies that characterizes function spaces explored by the convolutional layers or graph neural networks yet. Our empirical observation in \cref{fig:kde_MMD} suggests that in DC, there are cases where the mean of the latent feature distribution matches, but still differs in other statistics. This suggests that the feature extractors does not map the input to the universal RKHS. Therefore, a metric with stronger distribution matching power may be beneficial to the dataset condensation process.

\textbf{No penalization for outliers/extreme values. } The synthetic dataset is initialized by randomly choosing existing data records in the original dataset. Thus, it is probable that the initial synthetic dataset contains data records that yield extreme values in the latent space. During the distribution matching process, they may become more extreme, as the objective of MMD only matches the mean of the compared distributions. An example is shown in \cref{fig:kde_MMD}. In the synthetic dataset, the largest value for the shown latent feature is larger than the corresponding largest value in the original dataset. They provide little information and waste valuable memory space that is limited for the synthetic dataset. 
{
\setlength{\algomargin}{1.5em}
\begin{algorithm2e*}
\caption{Distribution Matching + Latent Quantile Matching}\label{alg:proposed}
\SetKwInOut{Input}{Input}
\SetKwInOut{Parameter}{Params}

\SetKwInOut{Output}{Output}

\Input{Training set $\mathcal{T}$}
\Parameter{Randomly initialized set of synthetic samples $\mathcal{S}$ for $\mathcal{C}$ classes, $|\mathcal{S}_c|=\beta_c$ for $c\in\mathcal{C}$, model $\phi_\theta$ parameterized with $\theta$, probability distribution over parameters $P_\theta$, train iterations $K$, learning rate $\eta$, quantile function $\mathcal{F}_q(Q,E):\mathbb{R}^{D\times F}\mapsto\mathbb{R}^{|Q|\times F}$, $Q$ is the set of quantiles, $E$ is the batch of embeddings with $D$ data records and $F$ features, sort function $\mathcal{F}_s(E):\mathbb{R}^{D\times F}\mapsto \mathbb{R}^{D\times F}$ sorts feature vectors $F$ in the ascending order.}

\For{$k = 0$ \KwTo $K-1$}{
Sample $\theta$ from $P_\theta$\\
\For{$c \in \mathcal{C}$}{
    Sample mini-batch $B^\mathcal{T}_{c}\sim\mathcal{T}$ and synthetic image $\mathcal{S}_c\subset\mathcal{S}$  for class $c$, $|S_c|=\beta_c$\\
    Compute the embeddings $\mathcal{E}^\mathcal{T}_c = \phi_\theta(x), x\in{B^\mathcal{T}_{c}}$ and $\mathcal{E}^\mathcal{S}_c = \phi_\theta(s), s\in\mathcal{S}_{c}$\\
    Compute the optimal quantiles $Q = \{\frac{2*1-1}{2\beta_c},\frac{2*2-1}{2\beta_c},...,\frac{2*\beta_c-1}{2\beta_c}\}$, $|Q|=\beta_c$\\
    
    Compute $\mathcal{L}=\frac{1}{\beta_c}||\mathcal{F}_q(Q,\mathcal{E}^\mathcal{T}_{c})-\mathcal{F}_s(\mathcal{E}^\mathcal{S}_{c})||^2$

    Update $S\gets S-\eta\Delta_S\mathcal{L}$
}
}
\Output{Condensed synthetic set $\mathcal{S}$}
\end{algorithm2e*}
}
\subsection{Proposed Method}
To address the abovementioned shortcomings, we propose our new method: Latent Quantile Matching (LQM) which learns a synthetic dataset that minimizes the mean squared error between the latent features in the synthetic dataset and the target quantiles of the latent feature distributions in the real dataset. The objective of LQM is to find the optimal synthetic dataset $S^*$ as defined in \cref{eqn:DM}, where $D$ denotes the sum of CvM test statistic (CvM stat) of the latent feature distributions. However, computing the CvM stat for each feature distribution for each epoch is time-consuming. Next, if we use it as the loss function, it will never reach zero. A zero CvM stat denotes that the ECDF of the synthetic latent distribution is the same as the ECDF of the real latent distribution, this is only possible if the number of data in the synthetic dataset is equal or larger than the number of data in the real dataset. Thus, a zero CvM stat is impossible and not desired in the scenario of DM based DC.

To establish an achievable objective, we utilize the optimal k-point discrete approximation of any distribution, as proposed in \citet{kennan2006note}. We define the set of optimal quantiles $Q$ that minimizes the CvM stats for a synthetic dataset with $k$ data records:
\begin{equation}\label{eqn:quantiles}
Q = \{\frac{2*1-1}{2k},\frac{2*2-1}{2k},...,\frac{2k-1}{2k}\}
\end{equation}
The ECDF constructed using values from these quantiles has the smallest CvM stat to the original distribution. The proof of this property can be found in \citet{barbiero2023discrete}. Using this property, and the fact that we have a predefined memory budget for the synthetic dataset, we can compute the quantiles that corresponds to the optimal points that minimizes the CvM stats for each latent feature distribution. This provides us with a realistic objective that can be reached for the condensation process: When the loss reaches zero, the synthetic latent distribution represents the most optimal approximation, measured by the CvM stat, of the real latent distribution when we only have $k$ data records. This objective is formally defined as:
\begin{equation}
\resizebox{0.91\columnwidth}{!}{
$S^* = \underset{S}{argmin}\frac{1}{|C|}\sum_{c\in C}||\mathcal{F}_{q}(Q,\phi_\theta(\mathcal{T}_c))-\mathcal{F}_{s}(\phi_\theta(\mathcal{S}_c))||^2$}
\end{equation}
Where $c\in C$, denotes the classes in the datasets. $Q$ denotes the optimal quantiles computed using \cref{eqn:quantiles}. $\mathcal{F}_q(Q,E):\mathbb{R}^{D\times F}\mapsto\mathbb{R}^{|Q|\times F}$ denotes a quantile computation function with a set of quantiles and a series of embeddings as inputs, and outputs the quantiles for each latent feature. $\mathcal{F}_s(E):\mathbb{R}^{D\times F}\mapsto \mathbb{R}^{D\times F}$ denotes a sort function that sorts each latent feature in the input embeddings in ascending order. LQM minimizes the distance between the latent feature in the synthetic dataset and the optimal quantiles that minimizes the CvM stat by sorting the synthetic latent feature and aligning them with the corresponding values of the optimal quantiles of the original latent feature distribution.

As our method only replaces the distance computation in the distribution matching process, it can be implemented on top of any existing DM based DC methods that compares the distance between two distributions. In \cref{alg:proposed}, the procedure of the proposed method, build on top of the basic distribution matching based DC algorithm is shown. 

\textbf{Limitation.} As the initial data records are selected randomly, when the memory budget becomes larger, the initial synthetic latent distribution will already be similar to the original latent distribution. This diminishes the advantage of LQM as the difference in high-order moments between the synthetic and real latent distributions is smaller. Therefore, the impact of the first shortcoming of MMD is smaller. Next, when the budget is large, the impact of few outliers being selected in the initialization process and kept in the synthetic dataset is less severe. There are many other data records with latent features within the normal range, which will provide the model with enough information. Thus, when the synthetic dataset is large, the improvement of LQM compared to MMD may be less noticeable.

\section{\uppercase{Experiments}}\label{section:experiments}

To validate the performance of LQM empirically. We implemented LQM on top of the SOTA DM based DC methods: `Improved Distribution Matching' (IDM) \cite{zhao2023improvedidm} and `Condense and Train' (CaT) \cite{liu2023catcgl} for image and graph-structured data, respectively. For DC with graph-structured data, we use the framework implemented by CaT-CGL \cite{liu2023catcgl} to evaluate our method in both the normal dataset condensation setting and in the continual graph learning (CGL) setting. CaT-CGL framework is implemented for CGL, However, the normal DC setting can be imitated by considering the full dataset as one task. We validate that LQM's  performance in CGL by implementing it in Cat-CGL, and compare it with CaT: a CGL method that utilizes DM based DC with MMD as distance function \cite{liu2023catcgl}. Additionally, we include two baselines for the CGL experiments: 1) Finetuning, 2) Joint. They refer to training a model in CGL setting without any CGL method, and training a model using the full dataset, respectively.

\textbf{Datasets.} For the evaluation with image data, we used datasets of various complexity and sizes. CIFAR-10, CIFAR-100 \cite{krizhevsky2009learningcifar} and TinyImageNet \cite{le2015tinyimagenet}. The task is to classify each image to the correct category. For graph-structured data, we used datasets from different domains. CoraFull \cite{bojchevski2017deepcorafull}, Arxiv \cite{hu2020openogbarxivproduct}, Reddit \cite{hamilton2017inductivereddit} and Product \cite{hu2020openogbarxivproduct} graph datasets were used in our experiments. The tasks of these datasets are to classify each node into their corresponding category. 

The details of the used datasets are reported in \cref{tab:imagedataset} and \cref{tab:graphdataset}. The budget per task row in \cref{tab:graphdataset} refers to the total number of nodes that we store in the synthetic dataset for each task in the CGL experiment. The budget is chosen such that each task is condensed to 1\% of the original size.

\begin{table}[h]
\label{tab:imagedataset} 

\begin{center}
\begin{small}
\begin{sc}
\begin{adjustbox}{width=\columnwidth,center}
\begin{tabular}{lccc}
\toprule
Dataset & CIFAR10 & CIFAR100 & TinyImageNet \\ 
\midrule
Images     & 60,000 &60,000&100,000\\ 
Image size & 32x32x3 & 32x32x3 &64x64x3\\ 
Classes    & 10      & 100 & 200\\ 
\bottomrule
\end{tabular}
\end{adjustbox}
\end{sc}
\end{small}
\end{center}
\vskip -0.1in
\caption{Statistics of the experimented image datasets.} 
\end{table}

\begin{table}[h]
\begin{center}
\label{tab:graphdataset} 
{\small{
\begin{adjustbox}{width=\columnwidth,center}
\begin{tabular}{lcccc}
\toprule
Dataset  & CoraFull & Arxiv & Reddit & Products \\ 
\midrule
Nodes    &19,793 &169,343 &227,853 &2,449,028\\
Edges    &130,622 &1,166,243&114,615,892&61,859,036\\
Features &8,710 &128 &602 &100\\
Classes &70&40&40&46\\
Tasks &35&20&20&23\\
\midrule
Budget/task & 4 & 29 & 40 & 219\\
\bottomrule
\end{tabular}
\end{adjustbox}
}}
\end{center}
\vskip -0.1in
\caption{Statistics of the experimented graph datasets.} 
\end{table}

\textbf{Experimental settings.} For both experiments, we evaluate the classification performance of deep learning models that are trained on the synthetic dataset learned by LQM. All experiments are run on a Tesla V100-SXM3-32GB GPU.

For image datasets, we learn 1/10/50 synthetic images per class. The training set is used to generate the synthetic dataset. The default hyperparameters of \citet{zhao2023improvedidm} are used for all of our experiments. The performance is measured by the mean accuracy and standard deviation of 5 runs, where the models are randomly initialized and trained on the condensed synthetic dataset and evaluated on the test set of the real dataset.

For graph datasets, we learn a synthetic dataset that is 1\% of the size of the original dataset. The number of nodes for each class is proportional to its ratio in the original dataset. In addition, a continual graph learning setting is used to show that LQM improves the performance in a more complex learning settings. Our work focus on the class-incremental CGL, where the model does not know from which task a sample belongs to. This is more difficult and realistic than the task-incremental setting, where the task identifier is available to the model at any time. To transform the dataset into the class-incremental setting, we first divided each dataset into train, validation, and test sets using a 6:2:2 ratio. Next, each set is divided into tasks. Each task only contains data of two unique, non-overlapping classes. The objective is to condense each task into a small synthetic dataset that is 1\% of the original size. We use the default hyperparameters of \citet{liu2023catcgl} for all of our experiments. Performance is measured by the mean and standard deviation of average accuracy (AA) and backward transfer (BWT) of 5 runs. In each run, a model sequentially learns different tasks on the condensed dataset.

The average accuracy is defined as the average of accuracies $A_{k,i}$ for each previously learned task $i$. It measures the average performance of the model across all learned tasks.
\begin{equation}
AA_k = \frac{1}{k}\sum^{k}_{i=1}A_{k,i}
\end{equation}
The backward transfer \cite{lopez2017gradient} indicates how the training of current tasks affects past tasks. A large number implies a greater impact, and the sign of the number denotes whether the impact is positive or negative. BWT is defined as:
\begin{equation}
BWT_k=\frac{1}{k-1}\sum^{k-1}_{i=1}(A_{k,i}-A_{i,i})
\end{equation}
\section{\uppercase{Results}}\label{section:results}
\subsection{Image datasets} 
We compare the performance of IDM+LQM against the SOTA dataset condensation (DC) methods in \cref{tab:imagepreformance} and the CvM test statistic (CvM stat) between the synthetic and the real dataset for IDM and IDM+LQM in \cref{fig:cvm}. We show the images in the synthetic dataset learned by IDM+LQM in the appendices. In \cref{tab:imagepreformance}, we include the performance of bi-level optimization based DC methods \cite{zhao2020datasetgradientmatching,zhao2021datasetaugmentation,wang2022cafe,lee2022dataset,du2023minimizing,cui2023scaling,guo2023towards} to show that distribution matching (DM) based DC methods can match their performance. However, the focus of our work is on the comparison between our method `IDM+LQM' and other DM based DC methods \cite{zhao2023improvedidm,sajedi2023datadamattention}. IDM+LQM outperforms both the reported and our replicated results of IDM on CIFAR-10 and CIFAR-100 for all synthetic dataset size. On TinyImageNet, IDM+LQM match the replicated performance on 1 and 10 image per class, and underperforms in the 50 image per class setting. 
\begin{table*}[h]

\begin{center}
\begin{small}
\begin{sc}
\begin{adjustbox}{width=\textwidth, center}
\begin{tabular}{c|c|ccc|ccc|ccc}
\toprule
Category& Methods& \multicolumn{3}{c|}{CIFAR-10} & \multicolumn{3}{c|}{CIFAR-100} & \multicolumn{3}{c}{TinyImageNet} \\
\midrule
\multicolumn{2}{c|}{Img / Cls}&1&10&50&1&10&50&1&10&50\\
\multicolumn{2}{c|}{Ratio(\%)}&0.02& 0.2& 1& 0.2& 2& 10& 0.2& 2& 10\\ 
\midrule
\multirow{8}{*}{Bi-level Optimization}&DC &
28.3±0.5 &44.9±0.5 &53.9±0.5 &12.8±0.3 &25.2±0.3 &-&-&-&-\\
&DSA &28.8±0.7 &52.1±0.5 &60.6±0.5 &13.9±0.3 &32.3±0.3 &42.8±0.4&-&-&-\\
&CAFE &30.3±1.1 &46.3±1.6 &55.5±0.6 &12.9±0.3 &27.8±0.3 &37.9±0.3&-&-&-\\
&CAFE+DSA &31.6±0.8 &50.9±0.5 &62.3±0.4 &14.0±0.3 &31.5±0.2 &42.9±0.2&-&-&-\\
&DSAC &34.0±0.7 &54.5±0.5 &64.2±0.4 &14.6±0.3 &33.5±0.3 &39.3±0.4&-&-&-\\
&FTD &46.8±0.3 &\underline{66.6±0.3} &73.8±0.2 &25.2±0.2 &43.4±0.3 &50.7±0.3&10.4±0.3&24.5±0.2&-\\
&TESLA &\underline{48.5±0.8} &66.4±0.8 &72.6±0.7 &24.8±0.4 &41.7±0.3 &47.9±0.3&-&-&-\\
&DATM &46.9±0.5 &\underline{66.8±0.2} &\underline{76.1±0.3} &\underline{27.9±0.2} &\underline{47.2±0.4} &\underline{55.0±0.2}&\underline{17.1±0.3}&\underline{31.1±0.3}&\underline{39.7±0.3}\\
\midrule
\midrule
\multirow{4}{*}{Distribution matching}&DM &26.0±0.8 &48.9±0.6 &63.0±0.4 &11.4±0.3 &29.7±0.3 &43.6±0.4 &3.9±0.2 &12.9±0.4 &24.1±0.3\\
&DataDAM &32.0±1.2 &54.2±0.8 &67.0±0.4 &14.5±0.5 &34.8±0.5 &49.4±0.3 &8.3±0.4 &18.7±0.3 &\textbf{28.7±0.3}\\
&IDM &45.6±0.7 &58.6±0.1 &67.5±0.1 &20.1±0.3 &45.1±0.1 &50.0±0.2 &10.1±0.2 &\textbf{21.9±0.2} &27.7±0.3\\
&IDM (replicated) &\textbf{45.3±0.5} &57.1±0.4 &67.2±0.2 &26.5±0.3 & 44.8±0.3 &50.5±0.2 &\textbf{10.3±0.4} &21.5±0.5 &25.2±0.6\\
\midrule
Full dataset & Full dataset &\multicolumn{3}{c|}{84.8±0.1} &\multicolumn{3}{c|}{56.2±0.3}&\multicolumn{3}{c}{ 37.6±0.4}\\
\midrule
\midrule
Ours&IDM+LQM &\textbf{45.9±0.6} &\textbf{60.9±0.3} &\textbf{70.2±0.1} &\textbf{27.2±0.4}&\textbf{47.7±0.3} &\textbf{52.4±0.4} &\textbf{10.4±0.3} &20.8±0.4 & 24.3±0.3\\
\bottomrule
\end{tabular}
\end{adjustbox}
\end{sc}
\end{small}
\end{center}  

\vskip -0.1in
    \caption{Comparison of test accuracy of randomly initialized model trained on condensed dataset. We evaluate our method on three different datasets with different numbers of synthetic images per class. Img/Cls: number of images per class. Ratio(\%): the ratio of condensed images to the whole training set. Full Dataset: the accuracy of the model trained on the whole training set. The bold results are the best performance of the distribution matching-based dataset condensation methods within the margin of error, excluding `Full Dataset' baseline. The best results of bi-level optimization based approaches are underlined. IDM+LQM uses the default parameter provided by IDM \protect\cite{zhao2023improvedidm}.}
\label{tab:imagepreformance}
\end{table*}
\begin{figure}[h]
\includegraphics[width=\columnwidth]{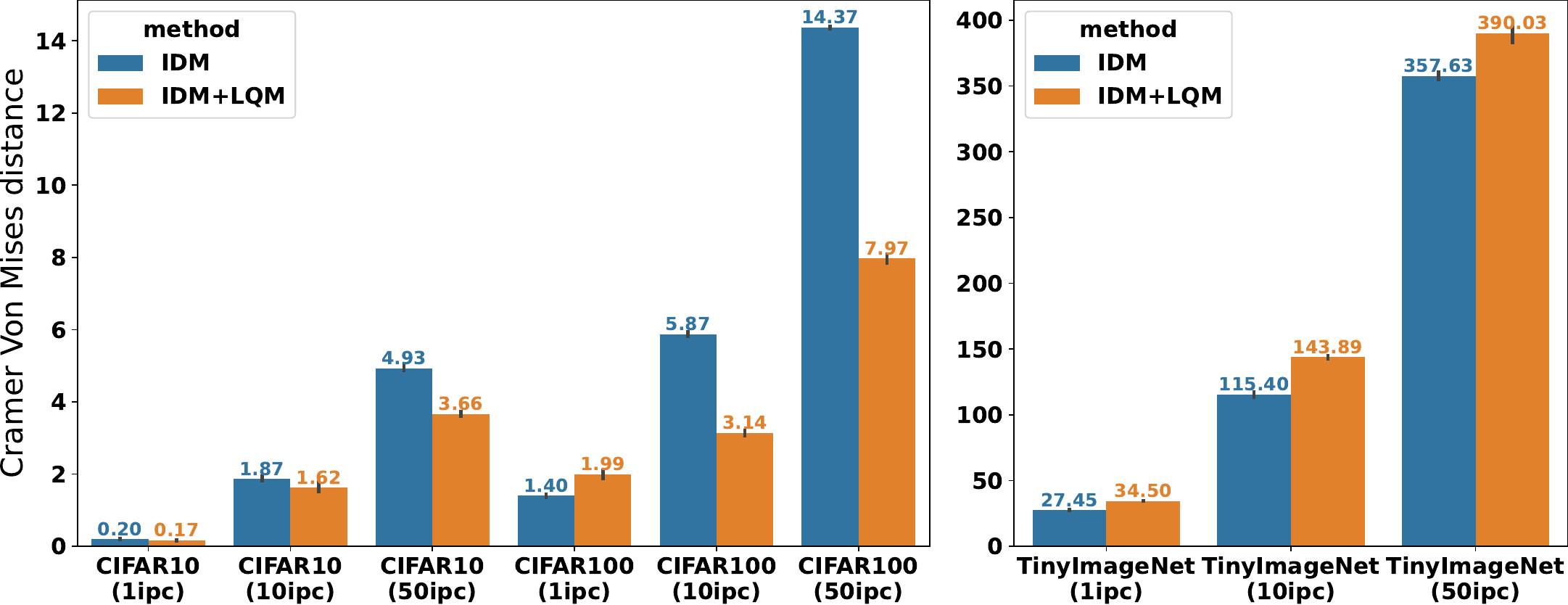}\label{fig:cvmcomp}
\centering
\vskip -0.05in
\caption{Comparison of average Cramér-von Mises stats between the synthetic dataset and the real dataset for each latent feature distribution. Lower number denotes higher probabilities that the compared samples (synthetic and real dataset in latent space) are drawn from the same distribution. The latent features are extracted by a pretrained model on the synthetic dataset.}
\label{fig:cvm}
\end{figure}

Comparing the results from \cref{tab:imagepreformance} and \cref{fig:cvm}, we see that IDM+LQM outperforms IDM when the CvM stat between the synthetic and real dataset in the latent space is lowered noticeably, i.e., with 10 and 50 image per class. This supports our hypothesis that using a distribution matching metric beyond the first moment yields better DC performance.

Next, we note that all our experiments are performed using the default hyperparameters provided by IDM, except the training epoch of the network on the synthetic dataset during the evaluation. We increased it from 1000 to 2000 to ensure the model learns the synthetic dataset produced by LQM optimally. As LQM changes the loss function of the method, the hyperparameters used by IDM with MMD may not be optimal to IDM with LQM anymore. Combined with the limitation we discussed in \cref{section:methodology}, our LQM underperforms on the TinyImageNet dataset. Due to the time constraint, we were unable to optimize the hyperparameters. However, we believe that the performance of IDM+LQM in CIFAR-10 and CIFAR-100 is already sufficient to demonstrate the efficacy of our new method.

\subsection{Graph datasets} 
To show that the efficacy of our method on complex learning settings, we compare the performance of CaT+LQM with other CGL baselines in \cref{tab:graphpreformance}. We experiment in two different settings, as indicated in the brackets on the category column. They are explained in \cref{section:experiments}. Our experiments use the default hyperparameters provided by CaT. Thus, the performance of CaT+LQM may benefit from further hyperparameter optimization. However, CaT+LQM already consistently matches or outperforms the replicated result of CaT that uses MMD. In the one task setting, CaT+LQM consistently outperform CaT. This validates that our method generalizes to graph-structured data. In the CGL setting, CaT+LQM has a higher average accuracy with similar or less negative impacts on past tasks in CGL setting, which is denoted by the higher backward transfer metric. 

We observe that the improvement of AA is larger when the synthetic dataset is small, corresponding to the limitation we discussed in \cref{section:methodology}. On CoraFull and Arxiv, we have 4 and 29 nodes for the synthetic dataset per task, respectively. Compared to the replicated results, LQM improves AA by $2\%$ and $1.4\%$, respectively. For larger datasets: Reddit and Product, where we have 40 and 318 nodes for the synthetic dataset per task, respectively, the improvement of LQM is less noticeable. LQM is $0.1\%$ worse and $0.7\%$ better in AA compared to the replicated results of CaT.

\begin{table*}[h]
\begin{center}
\begin{small}
\begin{sc}
\begin{adjustbox}{width=\textwidth, center}
\begin{tabular}{c|c|cc|cc|cc|cc}
\toprule
\multirow{2}{*}{Category} & \multirow{2}{*}{Methods} & \multicolumn{2}{c}{CoraFull} & \multicolumn{2}{c}{Arxiv} & \multicolumn{2}{c}{Reddit} & \multicolumn{2}{c}{Product}\\
\cmidrule{3-10}

&& AA(\%)$\uparrow$ & BWT(\%)$\uparrow$ &AA(\%)$\uparrow$&BWT(\%)$\uparrow$&AA(\%)$\uparrow$&BWT(\%)$\uparrow$&AA(\%)$\uparrow$&BWT(\%)$\uparrow$\\
\midrule
Dataset Condensation (One task)& CaT (replicated)&
54.6±1.0& - &64.8±1.3& - &92.7±0.3& - & 84.3±0.0 & -\\
\midrule
Ours (One task) & CaT+LQM&
\textbf{57.5±0.2}& -&\textbf{67.6±0.4}& - &\textbf{92.8±0.1}& - &\textbf{ 84.4±0.1 }& - \\
\midrule
\midrule
Lower Bound (CGL)& Finetuning&
2.2$\pm$0.0&-96.6$\pm$0.1&5.0$\pm$0.0&-96.7$\pm$0.1&5.0$\pm$0.0&-99.6$\pm$0.0&4.3$\pm$0.0&-97.2$\pm$0.1\\
\midrule
Dataset condensation& CaT & 64.5$\pm$1.4&\textbf{-3.3$\pm$2.6}&66.0$\pm$1.1&-13.1$\pm$1.0&\textbf{97.6}$\pm$0.1&\textbf{-0.2$\pm$0.2}&\textbf{71.0$\pm$0.2}&\textbf{-4.8$\pm$0.4}\\
(CGL)& CaT (replicated)&
66.1$\pm$0.6&-8.4$\pm$0.7&66.6$\pm$0.6&-12.4$\pm$0.5&97.2$\pm$0.0&-0.5$\pm$0.1&70.3$\pm$0.1&-5.0$\pm$0.1\\
\midrule

Full dataset (CGL)& Joint& 85.3$\pm$0.1&-2.7$\pm$0.0&63.5$\pm$0.3&-15.7$\pm$0.0&98.2$\pm$0.0&-0.5$\pm$0.0&72.2$\pm$0.4&-5.3$\pm$0.5\\

\midrule

Ours (CGL) & CaT+LQM&
\textbf{68.1$\pm$0.2}&-8.7$\pm$0.3&\textbf{68.0$\pm$0.3}&\textbf{-10.7$\pm$0.2}&97.1$\pm$0.0&-0.5$\pm$0.0&\textbf{71.0$\pm$0.2}&-4.9$\pm$0.2\\
\bottomrule

\end{tabular}
\end{adjustbox}

\end{sc}
\end{small}
\end{center}    

\vskip -0.1in
\caption{Comparison of AA and BWT of dataset condensation based CGL methods in Class-IL setting without inter-task edges. The bold results are the best performance excluding Joint baseline. ↑ denotes the greater value represents greater performance. Two different settings are expressed in the category column. CGL setting splits dataset into tasks with two classes per task, one task setting considers whole dataset as one task, which imitates the usual dataset condensation setting. In one task setting, BWT is not available as the model only learns one task, thus there is no transfer of knowledge between the tasks.}
\label{tab:graphpreformance}
\end{table*}
\begin{figure}[h]
\includegraphics[width=0.70\columnwidth]{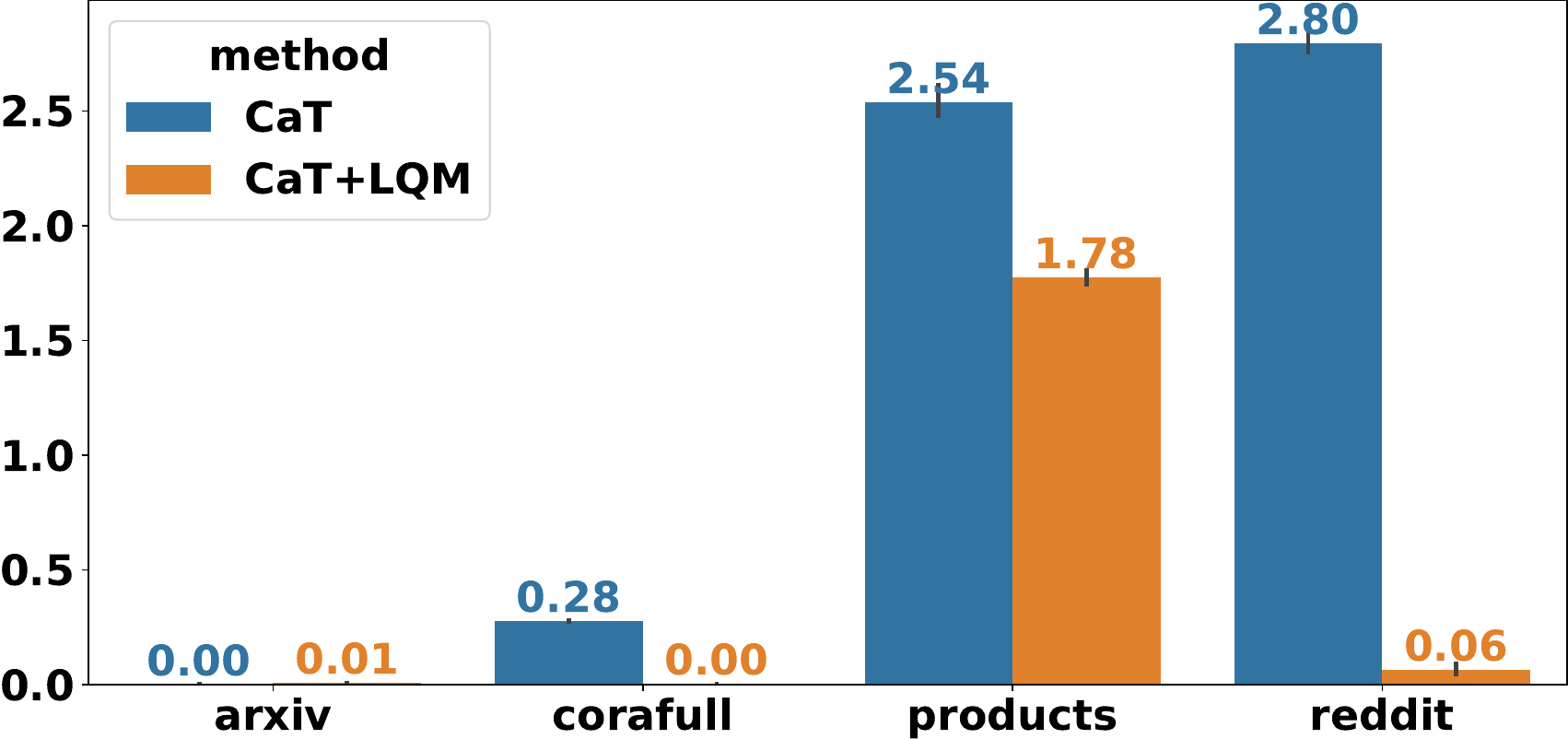}
\centering
\vskip -0.05in
\caption{The average percentage of extreme latent values for each class. A value of one denotes that one percent of the latent features in each class in the synthetic dataset exceed the maximum or dropped below the minimum of the corresponding class latent distribution in the real dataset. The latent features are extracted by a pretrained model on the synthetic dataset.}
\label{fig:extreme}
\end{figure}

The influence of the shortcomings of MMD is especially intriguing in the context of CGL experiments, where tasks are learned sequentially. When the first few tasks contain outliers, the model trained on these tasks will overfit to the outliers, which negatively impacts the generalization towards future tasks. In the non-continual learning experiments, all the data are available at the start, the model must generalize to all the samples at once. Thus, the effect of the outlier in the synthetic dataset is smaller. Our method penalizes the outliers explicitly, which results in better performance in the CGL setting. In \cref{fig:extreme}, we show the average number of latent features of each class in the synthetic datasets that is higher than the maximum or lower than the minimum of the corresponding latent feature in the real dataset of the same class. We observe that CaT+LQM has low(er) number of extreme values in the latent feature. However, this does not fully explain why Arxiv performs much better than Reddit. We observe that CoraFull and Arxiv both have a low number of edges per node. As the graph convolutional layer aggregates neighboring features, the extreme values will be smoothed out. When there are many neighbors, the majority of the neighbor features will not be extreme values. On CoraFull and Arxiv, the smoothing effect is not strong, thus, LQM outperforms the MMD counterpart.



\section{\uppercase{Conclusion}}\label{section:conclusion}
We propose a novel distribution matching (DM) based dataset condensation (DC) method: Latent Quantile Matching (LQM). It alleviates two shortcomings of the widely used Maximum Mean Discrepancy (MMD) based loss function, i.e., inadequate distribution matching power and lack of penalization for outliers/extreme values. Our method chooses specific values which the synthetic dataset should be aligned to based on the optimal k-point approximation that minimizes the Cramér-von Mises (CvM) test statistics between the two distributions, as studied in \citet{kennan2006note} and reported in \citet{barbiero2023discrete}. This alleviates the first shortcoming, as the goodness of fit tests matches the distributions on higher-order moments, while MMD only matches the mean, or the first-order moment. It also alleviates the second shortcoming, the extreme values will be penalized more as they are located further from the values at the optimal quantiles. 

As the core of our method is the adapted computation of the distance between two distributions, we can easily implement it on top of other DM based DC methods that use MMD \cite{zhao2023datasetdm,zhao2023improvedidm,liu2023catcgl,sajedi2023datadamattention}. Extensive empirical experiments on both image and graph datasets show that LQM outperforms or matches the performance of the counterpart that uses MMD. Moreover, the improvement of LQM is more noticeable with constraint memory budgets in the continual graph learning (CGL) setting. The severity of the two shortcomings increases as the memory budget decreases, since the outlier will have a higher influence within the limited synthetic dataset. This effect is more noticeable in CGL as the model can only learn a subset of the dataset for each task, therefore, the model is more likely to overfit to the outliers in the synthetic dataset. Thus, the improved result of LQM in CGL at low memory budget shows that LQM effectively alleviates the two shortcomings. As the objective of DC is to create small synthetic datasets, the advantage of LQM compared to MMD is going to be more prominent in future research. This property makes LQM a future direction on follow-up research regarding the DM distance functions in DM based DC methods.

Some future works include: 1) Evaluate the performance of LQM with other DM based DC methods. 2) Incorporate a heuristic based initialization procedure for the synthetic dataset to improve the performance. 3) Study the possibility of using other statistical metrics for DM based DC.

\section*{Acknowledgements}
This research received funding from the Flemish Government under the “Onderzoeksprogramma Artificiële Intelligentie (AI) Vlaanderen” programme.

{
    \small
    \bibliographystyle{ieeenat_fullname}
    \bibliography{main}
}


\newpage
\appendix
\onecolumn

\section{Ablation study}\label{section:ablation}
\textbf{Choices of goodness of fit tests. }
In \citet{barbiero2023discrete}, the optimal quantiles of the CvM and AD test statistics are reported. The optimal quantiles for AD test statistics can be obtained by an iterative process. We report the process proposed by \citet{barbiero2023discrete} in \cref{sec:ADalgo}. The quantiles obtained will minimize the AD test statistic, which is a variant of the CvM test statistic that gives more weight to the tails of the distribution. We have experimented with the optimal quantiles that minimize the AD test statistic with graph structured data. The result is shown in \cref{tab:ADperf}.

Compared to the performance of optimal CvM quantiles, the optimal AD quantiles does not provide noticeable improvements across the four experimented dataset. In the one task setting with graph structured data, the variant that uses optimal CvM quantiles consistently outperform the variant that uses optimal AD quantiles. Next, the computation of CvM quantiles does not require an iterative process. Thus, the use of the CvM test is more suitable for DM based DC.

\begin{table}[h]
    \label{tab:ADperf}
\begin{center}
\begin{small}
\begin{sc}

    \begin{adjustbox}{width=\textwidth, center}
    
    \begin{tabular}{c|c|cc|cc|cc|cc}
    \toprule
    \multirow{2}{*}{Setting} & \multirow{2}{*}{Methods} & \multicolumn{2}{c}{CoraFull} & \multicolumn{2}{c}{Arxiv} & \multicolumn{2}{c}{Reddit} & \multicolumn{2}{c}{Product}\\
    \cmidrule{3-10}
    
    &&AA(\%)$\uparrow$ & BWT(\%)$\uparrow$ &AA(\%)$\uparrow$&BWT(\%)$\uparrow$&AA(\%)$\uparrow$&BWT(\%)$\uparrow$&AA(\%)$\uparrow$&BWT(\%)$\uparrow$
    \\
    
    \midrule
    
    \multirow{2}{*}{One task}
    & CaT+LQM (AD)&
    57.0±0.2&- &67.2±0.2&- &\textbf{92.8±0.1}&- &84.3±0.1&-\\
    &CaT+LQM (CvM)&
    \textbf{57.5±0.2}& -&\textbf{67.6±0.4}& - &\textbf{92.8±0.1}& - &\textbf{84.4±0.1}& - \\

    \midrule
    \midrule

    \multirow{2}{*}{CGL}
    & CaT+LQM (AD)&
    \textbf{68.3$\pm$0.4}&\textbf{-8.7$\pm$0.2} &67.1$\pm$0.2&-11.8$\pm$0.6 &\textbf{97.2$\pm$0.1}&\textbf{-0.5$\pm$0.1} &\textbf{71.0$\pm$0.4}&-5.0$\pm$0.3\\
    &CaT+LQM (CvM)&
    68.1$\pm$0.2&\textbf{-8.7$\pm$0.3} & \textbf{68.0$\pm$0.3}&\textbf{-10.7$\pm$0.2} &97.1$\pm$0.0&\textbf{-0.5$\pm$0.0} &\textbf{71.0$\pm$0.2}&\textbf{-4.9$\pm$0.2} \\

    \bottomrule
    
    \end{tabular}
    \end{adjustbox}
    \end{sc}
    \end{small}
    \end{center}
    \caption{Comparison of AA of LQM using two different goodness of fit test statistic. The bold results are the best performance. ↑ denotes the greater value represents greater performance.}
\end{table}

\section{Algorithm of optimal quantiles for Anderson-Darling test statistic}\label{sec:ADalgo}
{
\setlength{\algomargin}{1.5em}
\begin{algorithm2e*}
    \caption{Optimal quantile computation for Anderson-Darling test statistic}\label{alg:ADdistance}
    \SetKwInOut{Input}{Input}
    \SetKwInOut{Parameter}{Params}

    \SetKwInOut{Output}{Output}

    \Input{Budget of discrete points $k$, epsilon $\epsilon_{max}$ for convergence checking}
    \Parameter{For $i=\{1,2,...,k\}$: quantile of the target distribution $q_i$, quantile of the discrete approximating distribution $Q_i$, probability of the discrete approximations $p_i$, loop counter $t$.}
  \BlankLine
    \BlankLine
    initialize $t = 1$\\
    \For{$i = 1$ \KwTo $k$}{
    initialize $p_{i}^0 = \frac{1}{k}, Q_{i}^0 = \frac{1}{k}.$
    }
    
    \While{$\epsilon_t > \epsilon_{max}$}{
        $t\leftarrow t+1$\\
        \For{$i = 1$ \KwTo $k$}{
            $q_{i}^t=\frac{Q_{i-1}^{t-1}+Q_{i}^{t-1}}{2}$\\
            $Q_{i}^t=log(\frac{1-q_i^t}{1-q_{i+1}^t})/log(\frac{q_{i+1}^t(1-q_i^t)}{q_i^t(1-q_{i+1}^t)})$\\
            $p_i^t=Q_i^t-Q_{i-1}^t$\\
        }
        $\epsilon_t = max_{i=1}^k|p_i^t-p_i^{t-1}|$\\
    }

   \Output{Quantiles $q_i$ that minimizes the Anderson-Darling test statistic.}
\end{algorithm2e*}
}

\section{Visualizations of synthetic image datasets learned by IDM+LQM}
\begin{figure}[h]
\includegraphics[width=0.58\textwidth]{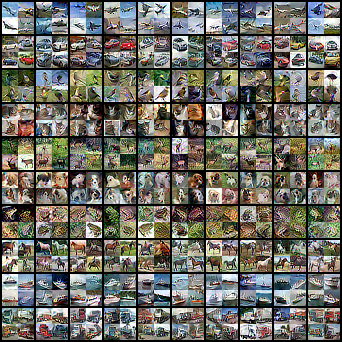}
\centering
\caption{Synthetic image dataset learned by IDM+LQM on CIFAR10 with 10 image per class, each row corresponds to a class.}
\label{fig:start}
\end{figure}

\begin{figure}[h]
\includegraphics[width=0.58\textwidth]{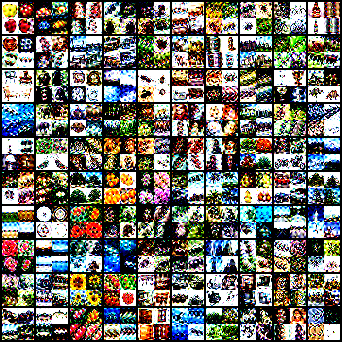}
\centering
\caption{Synthetic image dataset learned by IDM+LQM on CIFAR100 with 1 image per class.}
\label{fig:cifar1}
\end{figure}

\begin{figure}[h]
\includegraphics[width=0.58\textwidth]{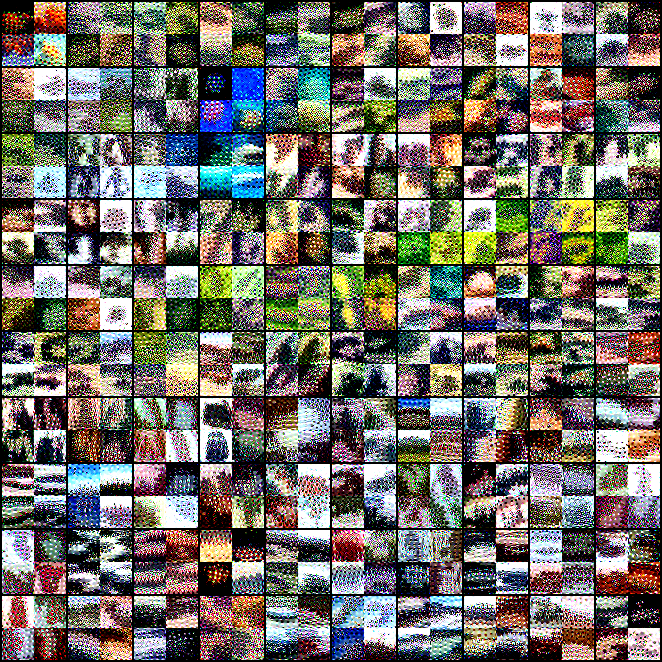}
\centering
\caption{Synthetic image dataset learned by IDM+LQM on TinyImageNet with 1 image per class. Only the first 100 classes are shown.}
\label{fig:tiny1}
\end{figure}

\begin{figure}[h]
\includegraphics[width=0.58\textwidth]{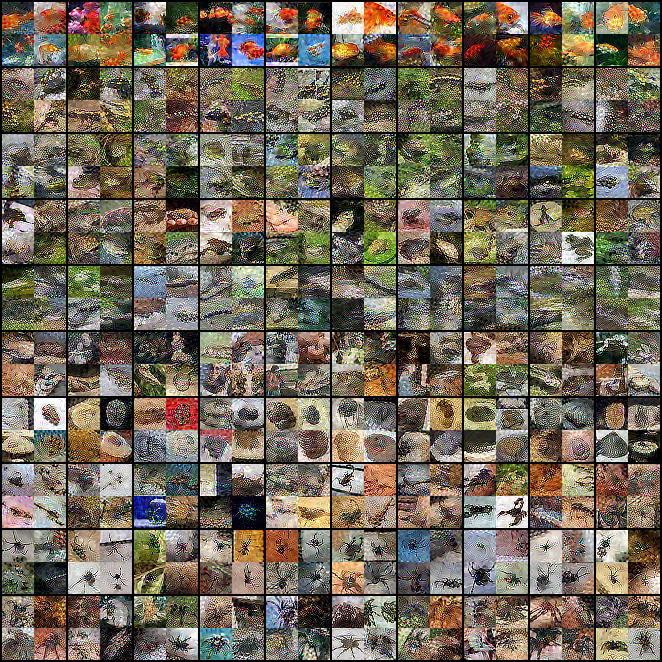}
\centering
\caption{Synthetic image dataset learned by IDM+LQM on TinyImageNet with 10 image per class. Only the first 10 classes are shown, each row corresponds to a class.}
\label{fig:end}
\end{figure}

We visualize the synthetic image dataset learned by IDM+LQM from \cref{fig:start} to \cref{fig:end}. We observe some repetitive dot patterns in the synthetic datasets learned in 1 image per setting, i.e., in \cref{fig:cifar1} and \cref{fig:tiny1}. In the corresponding 10 image per class setting for TinyImageNet demonstrated in \cref{fig:end}, this is less severe. This may indicate that in 1 image per class setting, the quantiles can't be matched perfectly if we want to maintain the image details.

\end{document}